\documentclass[letterpaper, 10 pt, conference]{ieeeconf}  
\usepackage{times}
\usepackage{graphicx}
\usepackage{amsmath}
\usepackage{amssymb}
\usepackage{subcaption}
\usepackage{multirow}
\usepackage{hyperref}

\IEEEoverridecommandlockouts                              
\def\FGPaperID{****} 

\title{\LARGE \bf
ElasticFace: Elastic Margin Loss for Deep Face Recognition}

\author{\parbox{16cm}{\centering
    {\large Huibert Kwakernaak$^1$ and Pradeep Misra$^2$}\\
    {\normalsize
    $^1$ Faculty of Electrical Engineering, Mathematics and Computer Science, University of Twente, Enschede, The Netherlands\\
    $^2$ Department of Electrical Engineering, Wright State University, Dayton, USA}}
    \thanks{This work was not supported by any organization}
}

\begin{document}

\author{Fadi Boutros$^{1,2}$, Naser Damer$^{1,2}$, 
Florian Kirchbuchner$^{1}$, Arjan Kuijper$^{1,2}$\\
$^{1}$Fraunhofer Institute for Computer Graphics Research IGD,
Darmstadt, Germany\\
$^{2}$Mathematical and Applied Visual Computing, TU Darmstadt,
Darmstadt, Germany\\
 Email: {fadi.boutros@igd.fraunhofer.de}
}
\pagestyle{plain}
\maketitle

\begin{abstract}
Learning discriminative face features plays a major role in building high-performing face recognition models. The recent state-of-the-art face recognition solutions proposed to incorporate a fixed penalty margin on commonly used classification loss function, softmax loss, in the normalized hypersphere to increase the discriminative power of face recognition models, by minimizing the intra-class variation and maximizing the inter-class variation. Marginal penalty softmax losses, such as ArcFace and CosFace, assume that the geodesic distance between and within the different identities can be equally learned using a fixed penalty margin. However, such a learning objective is not realistic for real data with inconsistent inter-and intra-class variation, which might limit the discriminative and generalizability of the face recognition model. In this paper, we relax the fixed penalty margin constrain by proposing elastic penalty margin loss (ElasticFace) that allows flexibility in the push for class separability. The main idea is to utilize random margin values drawn from a normal distribution in each training iteration. 
This aims at giving the decision boundary chances to extract and retract to allow space for flexible class separability learning. 
We demonstrate the superiority of our ElasticFace loss over ArcFace and CosFace losses, using the same geometric transformation, on a large set of mainstream benchmarks. From a wider perspective, our ElasticFace has advanced the state-of-the-art face recognition performance on seven out of nine mainstream benchmarks. All training codes, pre-trained models, training logs are publicly released \footnote{
\url{https://github.com/fdbtrs/ElasticFace}}.
\end{abstract}

\section{Introduction}


Face recognition technologies are increasingly deployed to enhance the security and convenience of processes involving identity verification, such as border control and financial services. The typical pipeline of a face recognition system involves mapping the face image (after detection and alignment \cite{zhang2016joint}) into a feature vector (embedding) \cite{deng2019arcface,DBLP:conf/cvpr/WangWZJGZL018,DBLP:conf/fgr/CaoSXPZ18}. Two face images are then compared by comparing their relative embeddings and therefore, measuring the degree of identity similarity between both faces. Knowing that it is intuitive that such embeddings should ideally have small intra-class and large inter-class variation, with the class here being an identity. This corresponds to a face recognition system that still makes correct genuine decisions (same identity) even when face images are largely varied (pose, age, expression, etc.), and make correct imposter (not same identity) decision even when the appearance of the face image pair of different identities is very similar. To achieve that, different solutions opted to train deep neural networks by either directly learning the embedding (e.g. Triplet loss \cite{DBLP:conf/cvpr/SchroffKP15}) or by learning an identity classification problem (e.g. Softmax loss \cite{DBLP:conf/fgr/CaoSXPZ18}).
One of the main challenges for training with metric-based learning such as Triple \cite{DBLP:conf/cvpr/SchroffKP15}, n-pair \cite{DBLP:conf/nips/Sohn16}, or contrastive \cite{DBLP:conf/cvpr/ChopraHL05} losses, is training the model with a large-scale dataset as the number of possible triplets explodes with the number of samples. Alternatively, classification-based losses such as softmax loss can be easily adopted for training a face recognition model as it does not pose that issue. However, the softmax loss does not directly optimize the feature embedding needed for face verification. 
Liu et al. \cite{DBLP:conf/icml/LiuWYY16} proposed a large-margin softmax (L-Softmax) by incorporating angular margin constraints on softmax loss to encourage intra-class compactness and inter-class separability between learned features. SphereFace \cite{DBLP:conf/cvpr/LiuWYLRS17} extended L-Softmax by normalizing the weights of the last full-connected layer and deploying multiplicative angular penalty margin between the deep features and their corresponding weights. 
Different from SphereFace, CosFace \cite{DBLP:conf/cvpr/WangWZJGZL018} proposed additive cosine margin on the cosine angle between the deep
features and their corresponding weights. CosFace also proposed to fix the norm of the deep features and their corresponding weights to 1, then scaling the deep feature norm to a constant $s$, achieving better performance on mainstream face recognition benchmarks. Later, ArcFace \cite{deng2019arcface}  proposed additive angular margin by deploying angular penalty margin on the angle between the deep features and their corresponding weights. The great success of softmax loss with penalty margin motivated several works to propose a novel variant of softmax loss \cite{dynarc,adaptiveface,uniformface,groupface,cricleloss,curricularface,magface,an2020partical_fc}. All these solutions achieved notable accuracies on mainstream benchmarks \cite{LFWTech,DBLP:conf/wacv/SenguptaCCPCJ16,DBLP:conf/cvpr/WhitelamTBMAMKJ17, DBLP:conf/icb/MazeADKMO0NACG18} for face recognition. 
Huang et al. \cite{curricularface} proposed an Adaptive Curriculum Learning loss based on margin-based softmax loss. The proposed loss targets the easy samples at an early stage of training and the hard ones at a later stage of training. Jiao et al. \cite{dynarc} proposed Dyn-arcface based on ArcFace loss \cite{deng2019arcface} by replacing the fixed margin value of ArcFace with an adaptive one. The margin value of Dyn-arcface is adjusted based on the distance between each class center and the other class centers. However, this might not reflect the real properties of the class separability, but rather their separability in the current stage of the model training.
Kim et al. \cite{groupface} proposed to enrich the feature representation learned by ArcFace loss with group-aware representations. UniformFace \cite{uniformface} suggested to equalize distances between all the classes centers by adding a new loss function to SphereFace loss \cite{DBLP:conf/cvpr/LiuWYLRS17}. A recent work by An et al. \cite{an2020partical_fc} presented an efficient distributed sampling algorithm (Partial-FC). 
The Partial-FC method is based on randomly sampling a small subset of the complete training set of classes for the softmax-based
loss function. Thus, it enables the training of the face recognition model on a massive number of identities. 
The authors experimentally proved that training with only 10\% of training samples using CosFace \cite{DBLP:conf/cvpr/WangWZJGZL018} and ArcFace\cite{deng2019arcface} can achieve comparable results on mainstream benchmarks to the case when training is performed on a complete set of classes.
MagFace \cite{magface} deployed magnitude-aware margin on ArcFace loss to enhance intra-class compactness by pulling high-quality
samples close to class centers while pushing low-quality samples away. However, this is based on the weak assumption of optimal face quality (utility) estimation. Moreover, this might prevent the model from convergence when the most of training samples in the training dataset are of low quality.

The main challenge for the majority of the previously listed works is the fine selection of the ideal margin penalty value.
In this work, we propose the ElasticFace loss that relaxes the fixed single margin value by deploying a random margin drawn from a normal distribution. 
We additionally extended this concept by guiding the assignment of the drawn margin values to put more attention on hardly classified samples.
We provided a simple toy example with an 8-class classification problem to demonstrate the enhanced separability and robustness induced by our ElasticFace loss.
To experimentally demonstrate the effect of our ElasticFace loss on face recognition accuracy, we report the results on nine different benchmarks. 
The achieved results are compared to the results reported in the recent state-of-the-art.
In a detailed comparison, compared to fixed margin penalties and recent state-of-the-art, our ElasticFace loss enhanced the face recognition accuracy on most of the considered benchmarks, consequently extending state-of-the-art face recognition performance on seven out of nine benchmarks and scoring close to the state-of-the-art in the remaining two. This is especially the case in the benchmarks where the intra-class variation is extremely high, such as frontal-to-profile face verification (CFP-FP \cite{DBLP:conf/wacv/SenguptaCCPCJ16}) and large age gap face verification (AgeDB-30 \cite{DBLP:conf/cvpr/MoschoglouPSDKZ17}), which points to the generalizability induced by the proposed ElasticFace.

In the rest of this paper, we will first introduce our proposed ElasticFace loss by building up to its definition starting from the basic softmax loss. This rationalization will include an experimental toy example demonstrating the effect of the proposed loss. Later on, the experimental setup and implementation details are introduced. This is followed by a detailed comparative discussion of the achieved results and a final conclusion.

\section{ElasticFace loss}
We propose in this work a novel learning loss strategy, ElasticFace loss, aiming at improving the accuracy of face recognition by targeting enhanced intra-class compactness and inter-class discrepancy in a flexible manner.
Unlike previous works \cite{deng2019arcface,DBLP:conf/cvpr/LiuWYLRS17,DBLP:conf/cvpr/WangWZJGZL018} that utilize a fixed penalty margin value, our proposed ElasticFace loss accommodates flexibility through relaxing this constraint by randomly drawing the margin value from a Gaussian distribution. 
Our proposed ElasticFace loss targets giving the model flexibility in optimizing the separability between and within the classes as it incorporates random margin values for each sample in each training iteration. 
The randomized margin penalty can be easily integrated into any of the angular margin-based softmax losses, which we demonstrate on two state-of-the-art margin-based softmax losses.
The angular margin-based losses and our ElasticFace loss extend over the softmax loss by manipulating the decision boundary to enhance intra-class compactness and inter-class discrepancy.
Therefore, in this section, we first revisit the conventional softmax loss.  Then, we present the modified version of softmax loss and the angular margin-based softmax loss. This leads up to presenting our proposed ElasticFace loss and an extended definition, the ElasticFace+, where the assignment of the drawn margins to training samples is linked to their proximity to their class centers.



\paragraph{Softmax Loss}
The widely used multi-class classification loss, softmax loss \cite{DBLP:conf/icml/LiuWYY16}, refers to applying cross-entropy loss between the output of the logistic function (softmax activation function) and the ground-truth.
Assume {$x_i \in R^d$} is a feature representation of the {i-th} sample {$z_i$} and {$y_i$} is its corresponding class label ($y_i$ integer in the range $[1,c]$). Given that {c} is the number of classes (identities),
the output of the softmax activation function is defined as follows:
\begin{equation}
    softmax(x_i,y_i)= \frac{e^{f_{y_i}}}{\sum\limits_{j=1}^{c}e^{f_j}}= \frac{e^{x_iW^T_{y_i}+b_{y_i}}  }{\sum\limits_{j=1}^{c} e^{x_iW^T_{j}+b_{j}}},
\end{equation}
where $f_{y_i}$ is the activation of the last fully-connected layer with weight vector $W_{y_i}$ and bias $b_{y_i}$.
$W_{y_i}$ is the $y_i$-th column of weights $W \in R^d_c$ and $b_{y_i}$ is the corresponding bias offset. The output of the softmax activation function is the probability of $x_i$ being correctly classified as $y_i$. Given a mini-batch of size N, the cross-entropy loss function that measures the  divergence between the model output and the ground-truth labels can be defined as follows:
\begin{equation}
    L_{CE}= \frac{1}{N} \sum\limits_{i \in N} -log \frac{e^{x_iW^T_{y_i}+b_{y_i}}  }{\sum\limits_{j=1}^{c} e^{x_iW^T_{j}+b_{j}}}.
\end{equation}
In a simple binary class classification, assuming that the input $z_i$ belong to class 1, the model will correctly classify $z_i$  if  $W_1^Tx_i+b1 > W_2^Tx_i+b2$ and $z_i$ will be classified as class 2 if $W_2^Tx_i+b2 > W_1^Tx_i+b1$. Therefore, the decision boundary of softmax loss is $x(W_1^T -W_2^T)+b1-b2=0$.
One of the main limitations of using softmax loss for learning face embeddings is that softmax loss does not explicitly optimize the feature representation needed for face verification as there is no restriction on the minimum distance between the class centers.
Thus, training with softmax loss is less than optimal for achieving the maximum inter-class distances and the minimum intra-class distances.
To mitigate this limitation,  margin penalty-based cosine softmax loss was proposed as an alternative to the conventional softmax loss and it became a popular loss function for training face recognition models \cite{deng2019arcface,DBLP:conf/cvpr/WangWZJGZL018,DBLP:conf/cvpr/LiuWYLRS17}. 
To get there, \cite{DBLP:conf/cvpr/LiuWYLRS17} has proposed a modified softmax loss (Cosine softmax loss) that optimized the angle cosine between features and the weights $cos(\theta)$ and then, incorporates a margin penalty on $cos(\theta)$.



\begin{figure*}[ht!]
     \centering
        \includegraphics[width=\textwidth]{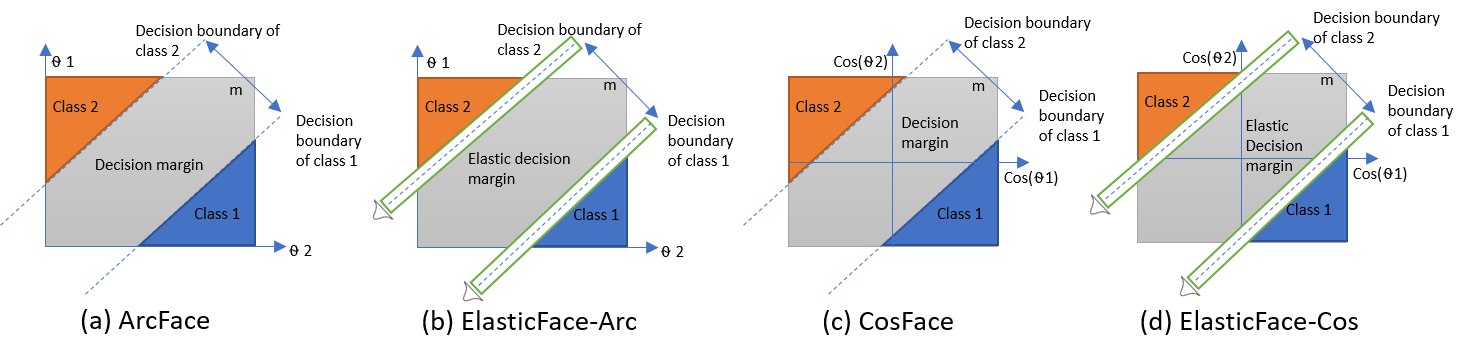}
    \caption{Decision boundary of (a) ArcFace, (b) ElasticFace-Arc, (c) CosFace, and (d) ElasticFace-Cos for binary classification. The dashed blue line is the decision boundary. The gray area illustrates the decision margin.}
    \label{fig:marg}
\end{figure*}
\paragraph{Cosine Softmax Loss}
Following \cite{deng2019arcface,DBLP:conf/cvpr/WangWZJGZL018,DBLP:conf/cvpr/LiuWYLRS17,DBLP:conf/icml/LiuWYY16}, the bias offset, for simplicity, can be fixed to $b_{y_i}=0$. The logit $f_{y_i}$, in this case, can be reformulated as:  ${x_iW^T_{y_i}}=\Vert x_i \Vert \Vert W_{y_i} \Vert cos(\theta_{y_i})$, where $\theta_{y_i}$ is the angle between the weights of the last fully-connected layer $W_{y_i}$ and the feature representation $x_i$. By fixing the weights norm and the feature norm to $\Vert W_{y_i} \Vert=1$ and $ \Vert x_i \Vert =1$, respectively, and rescaling the $\Vert x_i \Vert$ to the constant $s$ \cite{DBLP:conf/cvpr/WangWZJGZL018}, the output of the softmax activation function is subject to the cosine of the angle $\theta_{y_i}$.
The modified softmax loss ($L_{ML}$) can be defined, as  stated in \cite{DBLP:conf/cvpr/WangWZJGZL018,DBLP:conf/cvpr/LiuWYLRS17}, as follows:

\begin{equation}
    L_{ML}= \frac{1}{N} \sum\limits_{i \in N} - log \frac{e^{s (cos(\theta_{y_i}))}}{ e^{s  (cos(\theta_{y_i}))} +\sum\limits_{j=1 , j \ne y_i}^{c}  e^{s  (cos(\theta_{j}))}}.
\end{equation}
In the previous binary example, assume that the input $z_i$ belong to the class 1, $z_i$ will be correctly classified if $cos(\theta1)>cos(\theta2)$.
The decision boundary, in this case, is $cos(\theta1)-cos(\theta2)=0$.
Therefore, training with the modified (cosine) softmax loss emphasizes that the model prediction depends on the angle cosine between the features and the weights. 
However, and similar to conventional softmax loss, modified softmax loss does not explicitly optimize the feature representation needed for face verification.
This motivated the introduction of the angular margin penalty-based losses \cite{deng2019arcface,DBLP:conf/cvpr/WangWZJGZL018,DBLP:conf/cvpr/LiuWYLRS17}.

\paragraph{Angular Margin Penalty-based Loss}
In recent works  \cite{deng2019arcface,DBLP:conf/cvpr/WangWZJGZL018,DBLP:conf/cvpr/LiuWYLRS17}, different types of margin penalty are proposed to push the decision boundary of softmax, and thus enhance intra-class compactness and inter-class discrepancy aiming at improving the accuracy of face recognition.
The general angular margin penalty-based loss ($L_{AML}$) is defined as follows:

\begin{equation}
\resizebox{\linewidth}{!}{$
    L_{AML}=\frac{1}{N}  \sum\limits_{i \in N} - log \frac{e^{s (cos(m_1\theta_{y_i}+m_2)-m_3)}}{ e^{s(cos(m_1\theta_{y_i}+m_2)-m_3)} +\sum\limits_{j=1 , j \ne y_i}^{c}  e^{s ( cos(\theta_{j}))}},
$}\end{equation}
where $m_1$, $m_2$ and $m3$ are the margin penalty parameters proposed by SphereFace \cite{DBLP:conf/cvpr/LiuWYLRS17}, ArcFace \cite{deng2019arcface} and CosFace \cite{DBLP:conf/cvpr/WangWZJGZL018}, respectively.
In SphereFace \cite{DBLP:conf/cvpr/LiuWYLRS17}, multiplicative angular margin penalty is deployed by multiplying $\theta$ by $m_1=\alpha$ and setting $m_2=0$ and $m_3=0$ (  $\alpha>1.0$). The decision boundary of SphereFace is then $cos(m_1 \theta_{y_i})- cos(\theta_j)=0$.
Differently, CosFace \cite{DBLP:conf/cvpr/WangWZJGZL018} proposed additive cosine margin penalty by setting $m_1=1$, $m_2=0$ and $m_3=\alpha$ ($0<\alpha < 1-cos(\frac{\pi}{4})$). The decision boundary of CosFace is then $cos(\theta_{y_i})- cos(\theta_j) -m3=0$.
Later, ArcFace \cite{deng2019arcface} proposed additive angular margin penalty by setting up $m_1=1$, $m_2=\alpha$ and $m_3=0$ ($0<\alpha <1.0$). The decision boundary of ArcFace is then $cos(\theta_{y_i}+m2)- cos(\theta_j)=0$.

Even though, ArcFace \cite{deng2019arcface}, CosFace \cite{DBLP:conf/cvpr/WangWZJGZL018} and SphereFace \cite{DBLP:conf/cvpr/LiuWYLRS17} introduced the important concept of angular margin penalty on softmax loss, selecting a single optimal margin value ($\alpha$) is a critical issue in these works.
By setting up $m_1=1$, $m_2=0$ and $m_3=0$, ArcFace, CosFace and SphereFace are equivalent to the modified softmax loss. A reasonable choice could be selecting a large margin value that is close to the margin upper bound to enable higher separability between the classes.  However, when the margin value is too large, the model fails to converge, as stated in \cite{DBLP:conf/cvpr/WangWZJGZL018}.
ArcFace, CosFace, and SphereFace selected the margin value through trial and error assuming that the samples are equally distributed in geodesic space around the class centers. However, this assumption could not be held when there are largely different intra-class variations leading to less than optimal discriminative feature learning, especially when there are large variations between the samples/classes in the training dataset. This motivated us to propose ElasitcFace loss by utilizing random margin penalty values drawn from a Gaussian distribution aiming at giving the model space for flexible class separability learning.

\paragraph{Elastic Angular Margin Penalty-based Loss (ElasticFace)}
The proposed ElasticFace loss is extended over the angular margin penalty-based loss by deploying random margin penalty values drawn from a Gaussian distribution.
Formally, the probability density function of a normal distribution is defined as follows:
\begin{equation}
\label{eq:gaussian}
    f(x)=\frac{1}{\sigma \sqrt{2\pi}}e^{-\frac{1}{2}(\frac{x-\mu}{\sigma})^2},
\end{equation}
where $\mu$ is the mean of the distribution and $\sigma$ is its standard deviation. To demonstrate and prove our proposed elastic margin, we chose to integrate the randomized margin penalty in ArcFace (noted as ElasticFace-Arc) and CosFace (noted as ElasticFace-Cos) as they proved to have clearer geometric interpretation and achieved higher accuracy on mainstream benchmarks than the earlier SphereFace.
ElasticFace-Arc ($L_{EArc}$) can be defined as follows:
\begin{equation}
\resizebox{\linewidth}{!}{$
    L_{EArc}= \frac{1}{N}  \sum\limits_{i \in N} - log \frac{e^{s (cos(\theta_{y_i}+E(m,\sigma)))}}{ e^{s (cos(\theta_{y_i}+E(m,\sigma)))} +\sum\limits_{j=1 , j \ne y_i}^{c}  e^{s ( cos(\theta_{j}))}},$}
\end{equation}
and ElasticFace-Cos ($L_{ECos}$) can be defined as follows:
\begin{equation}
\resizebox{\linewidth}{!}{$
    L_{ECos}= \frac{1}{N} \sum\limits_{i \in N} - log \frac{e^{s (cos(\theta_{y_i})-E(m,\sigma))}}{ e^{s  (cos(\theta_{y_i})-E(m,\sigma))} +\sum\limits_{j=1 , j \ne y_i}^{c}  e^{s ( cos(\theta_{j}))}},$}
\end{equation}
where $E(m,\sigma)$ is a normal function that return a random value from a Gaussian distribution with the mean $m$ and the standard deviation $\sigma$.


The decision boundaries of ElasticFace-Arc and ElasticFace-Cos are $cos(\theta_{y_i}+E(m,\sigma))- cos(\theta_j)=0$ and $cos(\theta_{y_i})- cos(\theta_j)- E(m,\sigma)=0$, respectively.  Figure \ref{fig:marg} illustrates the decision boundary of ArcFace, ElasticFace-Arc, CosFace and ElasticFace-Cos.
The sample push towards its center during training using ElasticFace-Arc and ElasticFace-Cos varies between training samples, based on the margin penalty drawn from $E(m,\sigma)$. 
%
During the training phase, a new random margin is generated for each sample in each training iteration. This aims at giving the model flexibility in the push for class separability. When $\sigma$ is 0, our ElasticFace-Arc and ElasticFace-Cos are equivalent to ArcFace and CosFace, respectively.

\begin{figure*}[ht!]
     \centering
     \begin{subfigure}[t]{0.32\textwidth}
         \centering
         \includegraphics[width=\textwidth]{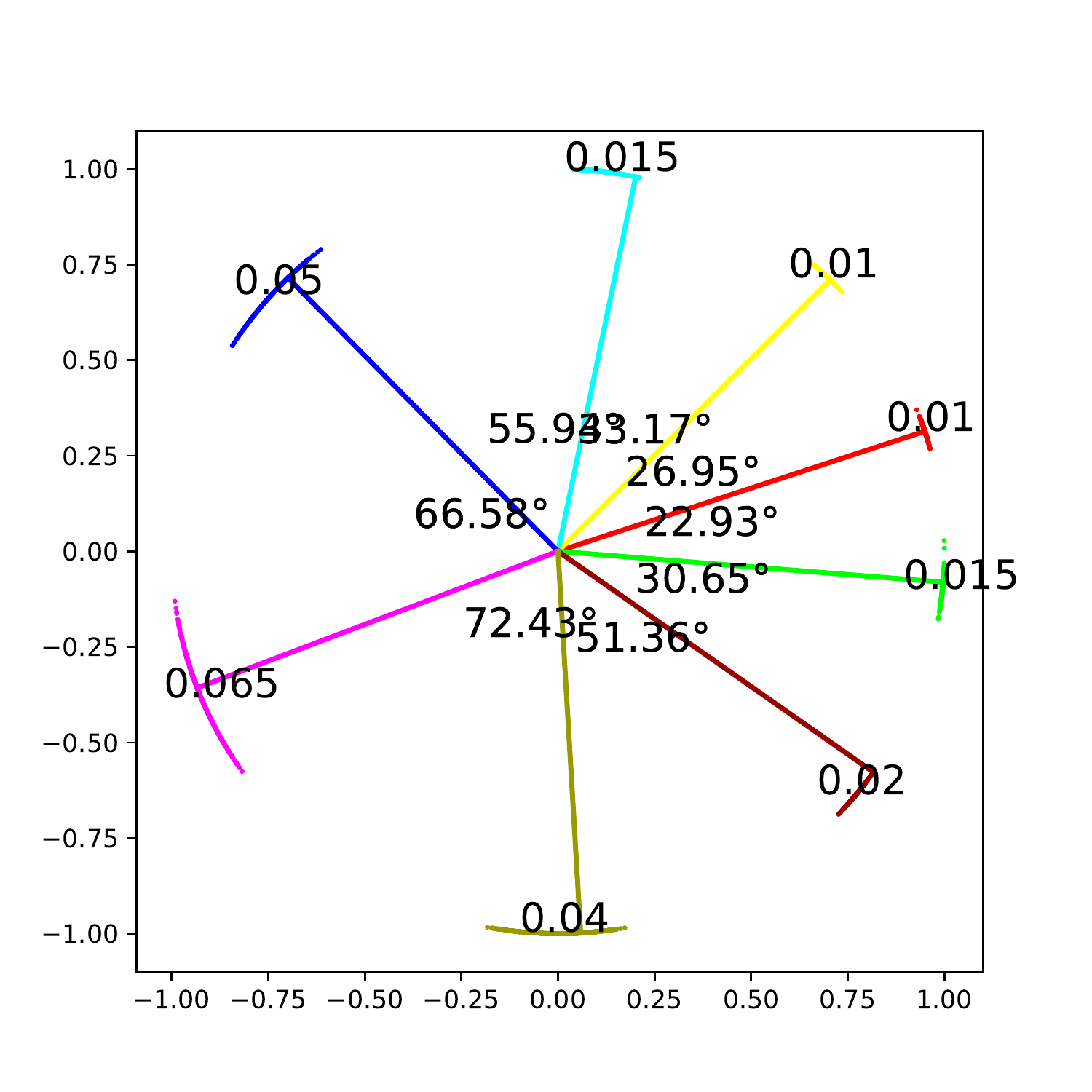}
         \caption{ArcFace ($m=0.5$)}
         \label{fig:AF5}
     \end{subfigure}
     \hfill
     \begin{subfigure}[t]{0.32\textwidth}
         \centering
         \includegraphics[width=\textwidth]{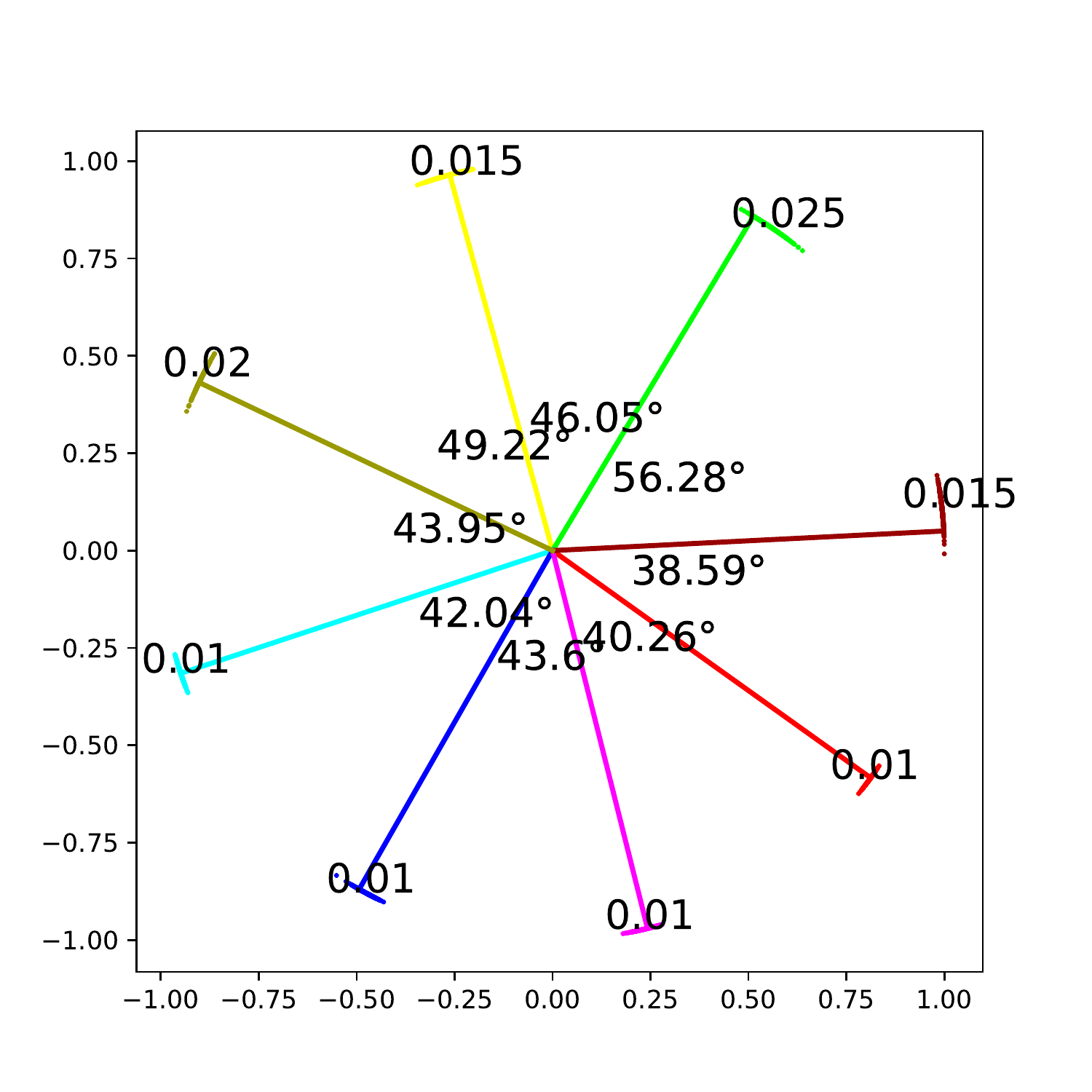}
         \caption{
         \centering
         ElasticFace-Arc ($m=0.5, \sigma=0.05$)}
         \label{fig:EF-A5}
     \end{subfigure}
     \hfill
     \begin{subfigure}[t]{0.32\textwidth}
         \centering
         \includegraphics[width=\textwidth]{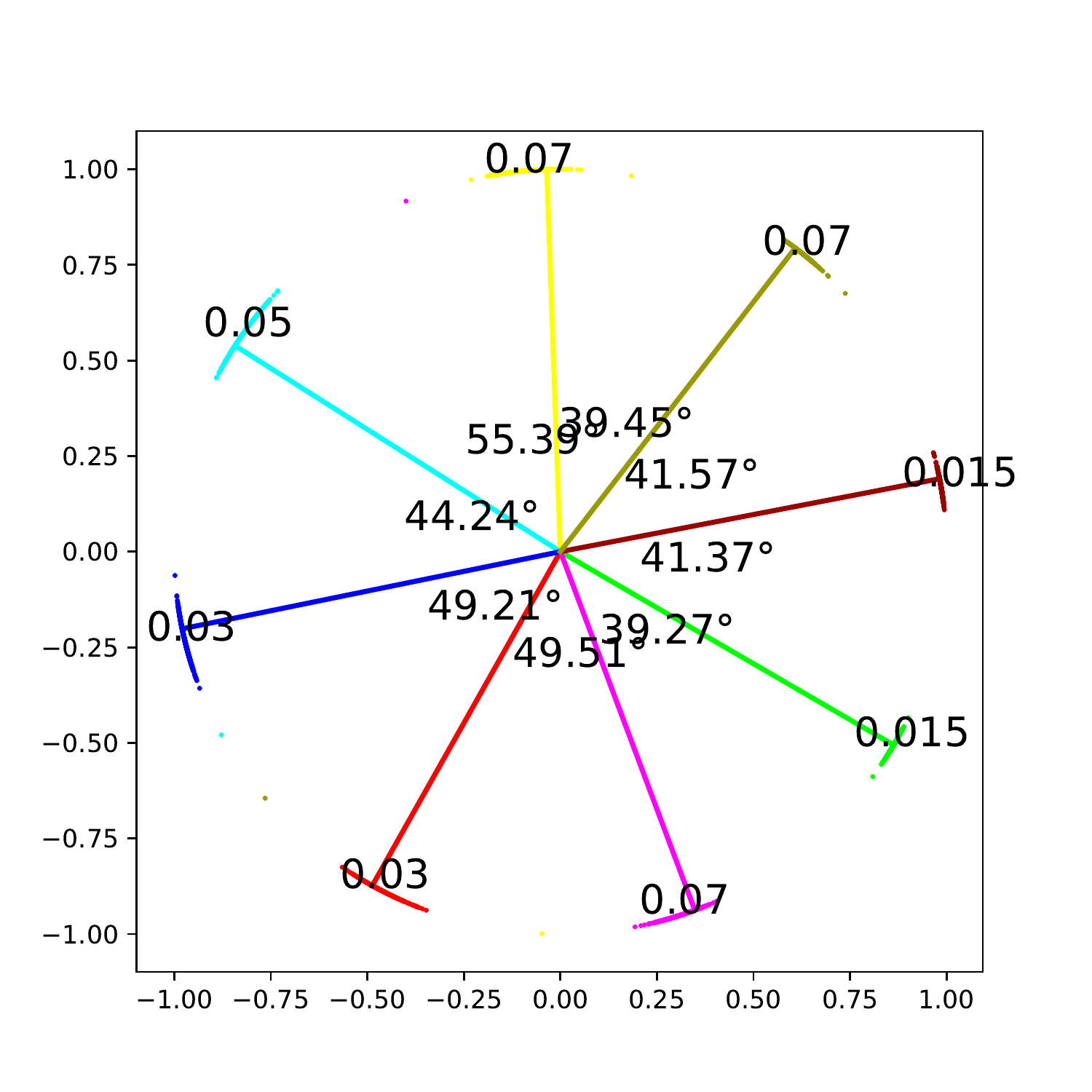}
         \caption{ElasticFace+ ($m=0.5, \sigma=0.0175$)}
         \label{fig:EF-A5Plus}
     \end{subfigure}
        \caption{Toy example of 3 ResNet-18 networks trained under different experimental settings. The 2-D features are normalized. Thus, the feature embeddings are allocated around the class centers in the arc space with a fixed radius.
        The numbers next to each class center indicate the mean of the standard deviation of each class feature embeddings. The angle in degree are calculated between each two consecutive classes to illustrate  the decision margin between the classes. 
        One can noticed that feature produced by ElasticFace and ElasticFace+ are more equally distributed around the class centers than ArcFace, in the arc space. Same colors always indicates same class across plots.
        }
        \label{fig:toy}
\end{figure*}

\paragraph{ElasticFace+}
We propose an extension to our ElasticFace, the ElasticFace+, that observes the intra-class variation during each training iteration and use this observation to assign a margin value to each sample based on its proximity to its class center.
This causes the samples that are relatively far from their class center to be pushed with a larger penalty margin to their class center.
This aims at giving the model space to push the samples that are relatively far from their class center to be closer to their centers while giving less penalty attention to the samples that are already close to their center.
To achieve that, we sort (descending) the output of the Gaussian distribution function (Equation \ref{eq:gaussian}) based on $cos(\theta_{y_i})$ value. 
Thus, the sample with small $cos(\theta_{y_i})$ will be pushed with large value from $E(m,\sigma)$ function, and vice versa. 

\begin{table}[ht!]
\centering
\resizebox{\linewidth}{!}{%
\begin{tabular}{|l|ll|ll|ll|ll|ll|l|}
\hline
\multirow{2}{*}{Loss} &
  \multicolumn{2}{l|}{LFW} &
  \multicolumn{2}{l|}{AgeDB-30} &
  \multicolumn{2}{l|}{CALFW} &
  \multicolumn{2}{l|}{CPLFW} &
  \multicolumn{2}{l|}{CFP-FP} &
  - \\ \cline{2-12} 
 &
  \multicolumn{1}{l|}{Acc (\%)} &
  BC &
  \multicolumn{1}{l|}{Acc(\%)} &
  BC &
  \multicolumn{1}{l|}{Acc(\%)} &
  BC &
  \multicolumn{1}{l|}{Acc(\%)} &
  BC &
  \multicolumn{1}{l|}{Acc(\%)} &
  BC &
  Sum BC \\ \hline
ArcFace (m=0.55) &
  \multicolumn{1}{l|}{\textbf{99.52 }} &
  3 &
  \multicolumn{1}{l|}{94.58} &
  1 &
  \multicolumn{1}{l|}{93.82} &
  2 &
  \multicolumn{1}{l|}{89.05} &
  1 &
  \multicolumn{1}{l|}{95.24} &
  1 &
  8 \\ \hline
ArcFace (m=0.5) &
  \multicolumn{1}{l|}{99.46} &
  2 &
  \multicolumn{1}{l|}{\textbf{94.83}} &
  3 &
  \multicolumn{1}{l|}{\textbf{93.88}} &
  3 &
  \multicolumn{1}{l|}{\textbf{89.72}} &
  3 &
  \multicolumn{1}{l|}{95.36} &
  2 &
  \textbf{13} \\ \hline
ArcFace(m=0.45) &
  \multicolumn{1}{l|}{99.43} &
  1 &
  \multicolumn{1}{l|}{94.66} &
  2 &
  \multicolumn{1}{l|}{93.80} &
  1 &
  \multicolumn{1}{l|}{89.42} &
  2 &
  \multicolumn{1}{l|}{\textbf{95.53}} &
  3 &
  9 \\ \hline \hline
ElasticFace-Arc(m=0.5, $\sigma$=0.0125) &
  \multicolumn{1}{l|}{\textbf{99.53}} &
  4 &
  \multicolumn{1}{l|}{94.80} &
  1 &
  \multicolumn{1}{l|}{93.68} &
  2 &
  \multicolumn{1}{l|}{89.72} &
  3 &
  \multicolumn{1}{l|}{95.43} &
  1 &
  11 \\ \hline
ElasticFace-Arc(m=0.5, $\sigma$=0.0175) &
  \multicolumn{1}{l|}{99.47} &
  1 &
  \multicolumn{1}{l|}{\textbf{95.13}} &
  4 &
  \multicolumn{1}{l|}{93.67} &
  1 &
  \multicolumn{1}{l|}{89.53} &
  2 &
  \multicolumn{1}{l|}{95.54} &
  3 &
  11 \\ \hline
ElasitcFace-Arc(m=0.5,$\sigma$=0.025) &
  \multicolumn{1}{l|}{99.52} &
  3 &
  \multicolumn{1}{l|}{94.95} &
  3 &
  \multicolumn{1}{l|}{93.78} &
  3 &
  \multicolumn{1}{l|}{89.50} &
  1 &
  \multicolumn{1}{l|}{95.44} &
  2 &
  12 \\ \hline
ElasitcFace-Arc(m=0.5,$\sigma$=0.05) &
  \multicolumn{1}{l|}{99.52} &
  3 &
  \multicolumn{1}{l|}{94.82} &
  2 &
  \multicolumn{1}{l|}{\textbf{93.90}} &
  4 &
  \multicolumn{1}{l|}{\textbf{89.79 }} &
  4 &
  \multicolumn{1}{l|}{\textbf{95.59}} &
  4 &
  \textbf{17} \\ \hline \hline
ElasitcFace-Arc+ (m=0.5,$\sigma$=0.0125) &
  \multicolumn{1}{l|}{\textbf{99.53}} &
  4 &
  \multicolumn{1}{l|}{95.00} &
  2 &
  \multicolumn{1}{l|}{93.68} &
  1 &
  \multicolumn{1}{l|}{\textbf{89.58}} &
  4 &
  \multicolumn{1}{l|}{95.40} &
  2 &
  13 \\ \hline
ElasitcFace-Arc+ (m=0.5, $\sigma$=0.0175) &
  \multicolumn{1}{l|}{\textbf{99.53}} &
  4 &
  \multicolumn{1}{l|}{95.07} &
  3 &
  \multicolumn{1}{l|}{93.95} &
  3 &
  \multicolumn{1}{l|}{89.37} &
  1 &
  \multicolumn{1}{l|}{\textbf{95.67}} &
  4 &
  \textbf{15} \\ \hline
ElasitcFace-Arc+ (m=0.5, $\sigma$=0.025) &
  \multicolumn{1}{l|}{99.42} &
  1 &
  \multicolumn{1}{l|}{\textbf{95.15}} &
  4 &
  \multicolumn{1}{l|}{93.73} &
  2 &
  \multicolumn{1}{l|}{89.48} &
  2 &
  \multicolumn{1}{l|}{95.36} &
  1 &
  10 \\ \hline
ElasitcFace-Arc+ (m=0.5,$\sigma$=0.05) &
  \multicolumn{1}{l|}{99.45} &
  2 &
  \multicolumn{1}{l|}{94.83} &
  1 &
  \multicolumn{1}{l|}{\textbf{94.00}} &
  4 &
  \multicolumn{1}{l|}{89.50} &
  3 &
  \multicolumn{1}{l|}{95.56} &
  3 &
  13 \\ \hline
\end{tabular}%
}
\caption{Parameter selection for ElasticFace-Arc and ElasticFace-Arc+. 
The Borda count (BC) is reported separately for each of training settings (ArcFace, ElasticFace-Arc and ElasticFace-Arc+)  and each of the evaluation benchmarks.  The final $\sigma$ and $m$ parameters are selected based on the highest BC sum.
In all settings, the used architecture is ResNet-50 trained on CASIA \cite{DBLP:journals/corr/YiLLL14a}.}
\label{tab:arc_param}
\end{table}

\begin{table}[ht!]
\centering
\resizebox{\linewidth}{!}{%
\begin{tabular}{|l|ll|ll|ll|ll|ll|l|}
\hline
\multirow{2}{*}{Loss} &
  \multicolumn{2}{l|}{LFW} &
  \multicolumn{2}{l|}{AgeDB-30} &
  \multicolumn{2}{l|}{CALFW} &
  \multicolumn{2}{l|}{CPLFW} &
  \multicolumn{2}{l|}{CFP-FP} &
  - \\ \cline{2-12} 
 &
  \multicolumn{1}{l|}{Acc (\%)} &
  BC &
  \multicolumn{1}{l|}{Acc (\%)} &
  BC &
  \multicolumn{1}{l|}{Acc (\%)} &
  BC &
  \multicolumn{1}{l|}{Acc (\%)} &
  BC &
  \multicolumn{1}{l|}{Acc (\%)} &
  BC &
  Sum BC \\ \hline
CosFace (m=0.4) &
  \multicolumn{1}{l|}{99.42} &
  1 &
  \multicolumn{1}{l|}{\textbf{94.65}} &
  3 &
  \multicolumn{1}{l|}{93.45} &
  1 &
  \multicolumn{1}{l|}{\textbf{90.38}} &
  3 &
  \multicolumn{1}{l|}{95.30} &
  1 &
  9 \\ \hline
CosFace (m=0.35) &
  \multicolumn{1}{l|}{\textbf{99.55}} &
  3 &
  \multicolumn{1}{l|}{94.55} &
  2 &
  \multicolumn{1}{l|}{\textbf{93.78}} &
  3 &
  \multicolumn{1}{l|}{89.95} &
  1 &
  \multicolumn{1}{l|}{95.31} &
  2 &
  \textbf{11} \\ \hline
CosFace (m=0.3) &
  \multicolumn{1}{l|}{99.45} &
  2 &
  \multicolumn{1}{l|}{94.45} &
  1 &
  \multicolumn{1}{l|}{93.46} &
  2 &
  \multicolumn{1}{l|}{90.12} &
  2 &
  \multicolumn{1}{l|}{\textbf{95.39}} &
  3 &
  10 \\ \hline \hline
ElasticFace-Cos (m=0.35,$\sigma$=0.0125) &
  \multicolumn{1}{l|}{99.45} &
  2 &
  \multicolumn{1}{l|}{94.72} &
  1 &
  \multicolumn{1}{l|}{93.83} &
  1 &
  \multicolumn{1}{l|}{90.12} &
  2 &
  \multicolumn{1}{l|}{95.47} &
  3 &
  9 \\ \hline
ElasticFace-Cos (m=0.35,$\sigma$=0.0175) &
  \multicolumn{1}{l|}{99.50} &
  3 &
  \multicolumn{1}{l|}{94.77} &
  3 &
  \multicolumn{1}{l|}{\textbf{93.97}} &
  4 &
  \multicolumn{1}{l|}{90.10} &
  1 &
  \multicolumn{1}{l|}{95.30} &
  2 &
  13 \\ \hline
ElasticFace-Cos (m=0.35,$\sigma$=0.025) &
  \multicolumn{1}{l|}{99.42} &
  1 &
  \multicolumn{1}{l|}{\textbf{94.85}} &
  4 &
  \multicolumn{1}{l|}{93.88} &
  2 &
  \multicolumn{1}{l|}{90.20} &
  3 &
  \multicolumn{1}{l|}{95.21} &
  1 &
  11 \\ \hline
ElasticFace-Cos (m=0.35,$\sigma$=0.05) &
  \multicolumn{1}{l|}{\textbf{99.52}} &
  4 &
  \multicolumn{1}{l|}{94.77} &
  3 &
  \multicolumn{1}{l|}{93.93} &
  3 &
  \multicolumn{1}{l|}{\textbf{90.38}} &
  4 &
  \multicolumn{1}{l|}{\textbf{95.52 }} &
  4 &
  \textbf{18} \\ \hline \hline
ElasticFace-Cos+(m=035, $\sigma$=0.0125 &
  \multicolumn{1}{l|}{99.38} &
  1 &
  \multicolumn{1}{l|}{94.50} &
  2 &
  \multicolumn{1}{l|}{93.67} &
  3 &
  \multicolumn{1}{l|}{89.85} &
  1 &
  \multicolumn{1}{l|}{95.20} &
  1 &
  8 \\ \hline
ElasticFace-Cos+(m=035, $\sigma$=0.0175) &
  \multicolumn{1}{l|}{99.45} &
  2 &
  \multicolumn{1}{l|}{\textbf{94.97}} &
  4 &
  \multicolumn{1}{l|}{93.48} &
  1 &
  \multicolumn{1}{l|}{89.98} &
  2 &
  \multicolumn{1}{l|}{95.23} &
  2 &
  11 \\ \hline
ElasticFace-Cos+(m=035, $\sigma$=0.025) &
  \multicolumn{1}{l|}{\textbf{99.55}} &
  4 &
  \multicolumn{1}{l|}{94.63} &
  3 &
  \multicolumn{1}{l|}{93.65} &
  2 &
  \multicolumn{1}{l|}{\textbf{90.28}} &
  4 &
  \multicolumn{1}{l|}{\textbf{95.47}} &
  4 &
  \textbf{17} \\ \hline
ElasticFace-Cos+(m=035, $\sigma$=0.05) &
  \multicolumn{1}{l|}{99.48} &
  3 &
  \multicolumn{1}{l|}{94.45} &
  1 &
  \multicolumn{1}{l|}{\textbf{93.77}} &
  4 &
  \multicolumn{1}{l|}{90.01} &
  3 &
  \multicolumn{1}{l|}{95.26} &
  3 &
  14 \\ \hline
\end{tabular}%
}
\caption{Parameter selection for ElasticFace-Cos and ElasticFace-Cos+. 
The Borda count (BC) is reported separately for each of training settings (ArcFace, ElasticFace-Cos and ElasticFace-Cos+) and each of the evaluation benchmarks.  The final $\sigma$ and $m$ parameters are selected based on the highest BC sum.
In all settings, the used architecture is ResNet-50 trained on CASIA \cite{DBLP:journals/corr/YiLLL14a}.}
\label{tab:cos_parameter}
\end{table}

\paragraph{Parameter Selection}
The probability density function has its peak around $m$ \cite{peebles1987probability}. 
Thus, when ElasticFace is integrated into ArcFace \cite{deng2019arcface}, we select the best margin value (as a single value) by training three instances of  ResNet-50 \cite{DBLP:conf/cvpr/HeZRS16} on CASIA \cite{DBLP:journals/corr/YiLLL14a} with ArcFace loss using margins equal to $0.45$, $0.50$ and $0.55$, respectively, to assure the advised margin in \cite{deng2019arcface}.
Then, based on the sum of the performance ranking Borda count on LFW \cite{LFWTech}, AgeDB-30  \cite{DBLP:conf/cvpr/MoschoglouPSDKZ17}, CALFW \cite{DBLP:journals/corr/abs-1708-08197}, CPLFW \cite{CPLFWTech}, and CFP-FP \cite{DBLP:conf/wacv/SenguptaCCPCJ16}, we select the margin that achieved the highest Borda count sum and set it as $m$ for $E(m,\sigma)$ function, where our goal is to use the most optimal margin. 
The best margin observed in our experiment, in this case, is 0.5 (Table \ref{tab:arc_param}). 
To select the $\sigma$ value for $E(m,\sigma)$ function, we conducted additional experiments on four instances of ResNet-50 trained on CASIA \cite{DBLP:journals/corr/YiLLL14a} with our proposed ElasticFace-Arc by setting up the $\sigma$ to one of these values $0.0125$, $0.015$, $0.025$ and $0.05$. 
Then, we rank these models based on the sum of the performance ranking Borda count across all datasets.
Finally, the $\sigma$ value is chosen based on the highest Borda count sum.  
The best $\sigma$ observed in our experiment, in this case, is 0.05 (Table \ref{tab:arc_param}). 
Similarly, we follow the same procedure to select the parameters ($m$ and $\sigma$) for ElasticFace-Cos.
We first choose the best margin value by training three different instances of ResNet-50 on CASIA \cite{DBLP:journals/corr/YiLLL14a} with CosFace using a margin equal to $0.3$, $0.35$, and $0.40$. 
The best $m$ observed in our experiment based on the sum of the performance ranking Borda count across all evaluated datasets, in this case, is 0.035 (Table \ref{tab:cos_parameter}). 
Similar to $\sigma$ selection approach of ElasticFace-Arc, we train four instance of ElasticFace-Cos to choose the best $\sigma$ for $E(m,\sigma)$ function. The best observed $\sigma$ in our experiment, in this case, is 0.05 (Table \ref{tab:cos_parameter}).
For ElasticFace-Cos+ and ElasticFace-Arc+, we followed the exact approach of parameter selection for ElasticFace-Arc and ElasticFace-Cos. The best observed $\sigma$ for ElasticFace-Arc+ is 0.0175 and the best observed one for ElasticFace-Cos+ is 0.025 (Table \ref{tab:arc_param} and \ref{tab:cos_parameter}).
These selected parameters are used to train our solutions (training details in Section \ref{sec:exp}) evaluated in Section \ref{sec:res}.


\paragraph{Toy example}
To demonstrate the robustness and the class separability induced by our proposed ElasticFace and ElasticFace+, we present a simple toy example by training three ResNet-18 networks \cite{DBLP:conf/cvpr/HeZRS16} to classify eight different identities and produce 2-D feature embeddings.  
All the networks are trained with a small batch size of 128 for 11200 iterations with stochastic gradient descent (SGD) and an initial learning rate of 0.1. The learning rate is reduced by a factor of 10 after 1680, 2800, 3360, and 8400 training iterations.
To demonstrate a classification case where the classes are not identically varied, these eight identities are selected to have four identities with small intra-class variation and another four identities with a large intra-class variation (measured as the average of all intra-class comparison scores for each identity).
These identities were chosen from all the identities with more than 400 images per identity in the MS1MV2 dataset \cite{deng2019arcface}, we note this selected subset as MS1MV2-400.
From these identities, we select the four identities with the highest intra-class variation and the four with the lowest intra-class variation. 
The features for this selection were extracted using an open-source \footnote {\url{https://github.com/deepinsight/insightface}} ResNet-100 \cite{DBLP:conf/cvpr/HeZRS16} model trained with ArcFace loss \cite{deng2019arcface}, and the comparison is performed by a cosine similarity. The set of the selected eight identities is noted as MS1MV2-8.
We use MS1MV2-8 to train the toy networks with ArcFace (m=0.5), ElasticArcFace (m=0.5, $\sigma$=0.05), and 
ElasticArcFace+ (m=0.5, $\sigma$=0.0175), based on our parameter selection.
Figure \ref{fig:toy} shows the classification of MS1MV2-8 for each of the experimental settings. In each of the plots in Figure
\ref{fig:AF5}, \ref{fig:EF-A5} and \ref{fig:EF-A5Plus}, we calculate the angle between each consecutive identities to demonstrate the separability between the identities in the arc space (inter-class discrepancy).  The optimal inter-class discrepancy may be achieved if the angle, in degree, between each of consecutive identities is close to 45 degrees i.e. 360 / 8. Also, we calculate the mean of the standard deviation of each class feature embeddings to illustrate intra-class compactness induced by ArcFace, ElasticFace, and ElasticFace+. The smaller standard deviation (shown at the edge of each class in Figure \ref{fig:toy}), in this case, indicates higher intra-class compactness.
It can be noticed that our EalsticFace and EalsticFace+ achieved better intra-class compactness and inter-class discrepancy than ArcFace, while the differences in inter-class variation between  EalsticFace and EalsticFace+ are minor (Figures \ref{fig:AF5} \ref{fig:EF-A5Plus}, and  \ref{fig:EF-A5}).












\begin{table*}[ht!]
\resizebox{\linewidth}{!}{%
\begin{tabular}{|l|c|c|c|c|c|c|}
\hline
\multirow{2}{*}{Method} & \multirow{2}{*}{\begin{tabular}[c]{@{}c@{}}Training \\ Dataset\end{tabular}} & LFW            & AgeDB-30       & CALFW         & CPLFW          & CFP-FP         \\ \cline{3-7} 
                        &                                                                              & Accuracy (\%)  & Accuracy (\%)  & Accuracy (\%) & Accuracy (\%)  & Accuracy (\%)  \\ \hline
ArcFace\cite{deng2019arcface} (CVPR2019)                 & MS1MV2  \cite{DBLP:conf/eccv/GuoZHHG16,deng2019arcface}                                                                     & 99.82  (3)        &          98.15     & 95.45         & 92.08          &98.27             \\
CosFace\cite{DBLP:conf/cvpr/WangWZJGZL018} (CVPR2018)                 & private                                                                      & 99.73          & -              & -             & -              & -              \\
Dynamic-AdaCos\cite{adacos} (CVPR2019)         & clean MS1M \cite{DBLP:conf/eccv/GuoZHHG16,adacos} + 
CASIA \cite{DBLP:journals/corr/YiLLL14a}                                                    & 99.73          & -              & -             & -              & -              \\
AdaptiveFace\cite{adaptiveface} (CVPR2019)            & clean MS1M    \cite{DBLP:conf/eccv/GuoZHHG16,DBLP:journals/tifs/WuHST18}                                                                   & 99.62          & -              & -             & -              & -              \\
UniformFace\cite{uniformface} (CVPR2019)            & clean MS1M \cite{DBLP:conf/eccv/GuoZHHG16,deng2019arcface}   + VGGFace2  \cite{DBLP:conf/fgr/CaoSXPZ18}                                                                & 99.8           & -              & -             & -              & -              \\
GroupFace\cite{groupface} (CVPR2020)              &  clean MS1M \cite{DBLP:conf/eccv/GuoZHHG16,deng2019arcface}                                                                  & \textbf{99.85} (1) & 98.28    (3)      & \textbf{96.20} (1) & 93.17 & 98.63          \\
CircleLoss\cite{cricleloss} (CVPR2020)            & clean MS1M \cite{DBLP:conf/eccv/GuoZHHG16,cricleloss}                                                           & 99.73          & -              & -             & -              & 96.02          \\
CurricularFace\cite{curricularface} (CVPR2020)          & MS1MV2  \cite{DBLP:conf/eccv/GuoZHHG16,deng2019arcface}                                                                     & 99.80           & 98.32 (2)         & \textbf{96.20} (1) & 93.13          & 98.37          \\
Dyn-arcFace \cite{dynarc} (MTAP2021)            & clean MS1M \cite{DBLP:conf/eccv/GuoZHHG16,deng2019arcface}                                                                        & 99.80           & 97.76          & -             & -              & 94.25          \\
MagFace\cite{magface} (CVPR2021)                 & MS1MV2  \cite{DBLP:conf/eccv/GuoZHHG16,deng2019arcface}                                                                      & 99.83    (2)      & 98.17          & 96.15         & 92.87          & 98.46          \\
Partial-FC-ArcFace \cite{an2020partical_fc} (ICCVW2021)    & MS1MV2 \cite{DBLP:conf/eccv/GuoZHHG16,deng2019arcface}                                                                      & 99.83 (2)         & 98.20          & 96.18   (2)      & 93.00             & 98.45          \\
Partial-FC-CosFace \cite{an2020partical_fc} (ICCVW2021)   & MS1MV2  \cite{DBLP:conf/eccv/GuoZHHG16,deng2019arcface}                                                                     & 99.83  (2)        & 98.03          & \textbf{96.20} (1) & 93.10           & 98.51          \\ \hline
ElasticFace-Arc (ours)    & MS1MV2 \cite{DBLP:conf/eccv/GuoZHHG16,deng2019arcface}                                                                      & 99.80          & \textbf{98.35} (1) & 96.17 (3)        & 93.27   (2)       & 98.67 (2) \\
ElasticFace-Cos (ours)  & MS1MV2   \cite{DBLP:conf/eccv/GuoZHHG16,deng2019arcface}                                                                    & 99.82   (3)       & 98.27      & 96.03         & 93.17 & 98.61  (3)      \\  
ElasticFace-Arc+ (ours)    & MS1MV2 \cite{DBLP:conf/eccv/GuoZHHG16,deng2019arcface}                                                                      & 99.82   (3)       & \textbf{98.35} (1) & 96.17 (3)         & \textbf{93.28} (1)       & 98.60 \\
ElasticFace-Cos+ (ours)  & MS1MV2   \cite{DBLP:conf/eccv/GuoZHHG16,deng2019arcface}                                                                    & 99.80          & 98.28   (3)       & 96.18 (2)        & 93.23 (3) &\textbf{98.73}  (1)        \\ \hline
\end{tabular}
}
\vspace{2mm}
\caption{The achieved results on the LFW, AgeDB-30, CALFW, CPLFW, and CFP-FP benchmarks. On large age gape (AgeDB-30) and frontal-to-profile face comparisons (CFP-FP), the ElasticFace solutions consistently extend state-of-the-art performances. ElasticFace scores very close to the state-of-the-art on LFW and CALFW. All decimal points are provided as reported in the respective works. The top performance in each benchmark is in bold. The top three performances in each benchmark are noted with rank number between parentheses (1,2 or 3).  }
\label{tab:res_lfw}
\end{table*}

\section{Experimental setup}
\label{sec:exp}
\paragraph{Training settings:}
The network architecture we used to demonstrate our ElasticFace is the ReseNet-100 \cite{DBLP:conf/cvpr/HeZRS16}. This was motivated by the wide use of this architecture in the state-of-the-art face recognition solutions \cite{deng2019arcface,an2020partical_fc,uniformface,cricleloss,curricularface}. We follow the common setting \cite{deng2019arcface,an2020partical_fc,curricularface} to set the scale parameter $s$ to 64.  
We set the mini-batch size to 512 and train our model on one Linux machine (Ubuntu 20.04.2 LTS) with Intel(R) Xeon(R) Gold 5218 CPU  2.30GHz, 512 G RAM, and  4 Nvidia GeForce RTX 6000 GPUs.
The proposed models in this paper are implemented using Pytorch \cite{NEURIPS2019_9015}.
All models are trained with Stochastic Gradient Descent (SGD) optimizer with an initial learning rate of 1e-1. 
We set the momentum to 0.9 and the weight decay to 5e-4. The learning rate is divided by 10 at 80k, 140k, 210k, and 280k training iterations. The total number of training iteration is 295K, which corresponds to the number of margin sampling from the normal distribution. 
During the training, we use random horizontal flipping with a probability of 0.5 for data augmentation.
The networks are trained (and evaluated) on images of the size $112 \times 112 \times 3$ to produce $512-d$ feature embeddings. 
These images are aligned and cropped using the Multi-task Cascaded Convolutional Networks (MTCNN)  \cite{zhang2016joint} following \cite{deng2019arcface}. All the training and testing images are normalized to have pixel values between -1 and 1.

\paragraph{Training dataset:} We follow the trend in recent works \cite{deng2019arcface,an2020partical_fc,curricularface,magface} in using the MS1MV2 dataset \cite{deng2019arcface} to train the investigated models with the proposed ElasticFace loss. 
This enables a direct comparison with the state-of-the-art as will be shown in Section \ref{sec:res}.
The MS1MV2 is a refined version \cite{deng2019arcface} of the MS-Celeb-1M \cite{DBLP:conf/eccv/GuoZHHG16} containing 5.8M images of 85K identities. 




\paragraph{Evaluation benchmarks and metrics:}

To demonstrate the effect of our proposed ElasticFace on face recognition accuracy and enable a wide comparison to state-of-the-art, we report the achieved results on nine benchmarks. 
These benchmarks are of a diverse nature, where some represent a special vulnerabilities of face recognition. The nine benchmarks are 1) Labeled Faces in the Wild (LFW) \cite{LFWTech}, 2) AgeDB-30 \cite{DBLP:conf/cvpr/MoschoglouPSDKZ17}, 3) Cross-age LFW (CALFW) \cite{DBLP:journals/corr/abs-1708-08197}, 4) Cross-Pose LFW (CPLFW) \cite{CPLFWTech}, 5) Celebrities in Frontal-Profile in the Wild (CFP-FP) \cite{DBLP:conf/wacv/SenguptaCCPCJ16}, 6) IARPA Janus Benchmark-B  (IJB-B) \cite{DBLP:conf/cvpr/WhitelamTBMAMKJ17}, 7) IARPA Janus Benchmark-C (IJB-C) \cite{DBLP:conf/icb/MazeADKMO0NACG18}, 8) MegaFace \cite{DBLP:conf/cvpr/Kemelmacher-Shlizerman16}, and 9) MegaFace (R) \cite{deng2019arcface}. 
The face recognition performance on LFW, AgeDB-30, CALFW, CPLFW, and CFP-FP is reported as verification accuracy, following their evaluation protocol. 
The performance on IJB-C and IJB-B is reported (as defined in \cite{DBLP:conf/cvpr/WhitelamTBMAMKJ17,DBLP:conf/icb/MazeADKMO0NACG18}) as true acceptance rates (TAR) at false acceptance rates (FAR) of 1e-4. 
The MegaFace and MegaFace(R) benchmarks report the face recognition performance as Rank-1 correct identification rate and as TAR at FAR=1e–6 verification accuracy.

We acknowledge the verification and identification performance evaluation metrics reported in ISO/IEC 19795-1 \cite{iso_metric}. However, to enhance the reproducibility and comparability, we follow the evaluation protocols and metrics used in each of the benchmarks as listed above. 

\begin{table*}[ht!]
\resizebox{\linewidth}{!}{%
\begin{tabular}{|l|c|c|c|c|c|c|c|}
\hline
\multirow{2}{*}{Method} & \multirow{2}{*}{\begin{tabular}[c]{@{}c@{}}Training\\ Dataset\end{tabular}} & IJB-B          & IJB-C          & \multicolumn{2}{c|}{MegaFace (R)}                                               & \multicolumn{2}{c|}{MegaFace}                                                   \\ \cline{3-8} 
                        &                                                                             & \begin{tabular}[c]{@{}c@{}}TAR at \\ FAR1e–4 (\%)\end{tabular}       & \begin{tabular}[c]{@{}c@{}}TAR at \\ FAR1e–4 (\%)\end{tabular}       & Rank-1 (\%)    & \begin{tabular}[c]{@{}c@{}}TAR at \\ FAR1e–6 (\%)\end{tabular} & Rank-1 (\%)    & \begin{tabular}[c]{@{}c@{}}TAR at \\ FAR1e–6 (\%)\end{tabular} \\ \hline
ArcFace\cite{deng2019arcface} (CVPR2019)                 & MS1MV2 \cite{DBLP:conf/eccv/GuoZHHG16,deng2019arcface}                                                                     & 94.2           & 95.6           & 98.35          & 98.48                                                          & 81.03          & 96.98                                                          \\
CosFace\cite{DBLP:conf/cvpr/WangWZJGZL018} (CVPR2018)                 & private                                                                     & -              & -              & -              & -                                                              & \textbf{82.72} (1) & 96.65                                                          \\
Dynamic-AdaCos\cite{adacos} (CVPR2019)          & clean MS1M \cite{DBLP:conf/eccv/GuoZHHG16,adacos} + 
CASIA \cite{DBLP:journals/corr/YiLLL14a}                                                                                               & -              & 92.40           & 97.41              & -                                                              & -              & -                                                              \\
AdaptiveFace\cite{adaptiveface} (CVPR2019)           & clean MS1M    \cite{DBLP:conf/eccv/GuoZHHG16,DBLP:journals/tifs/WuHST18}                                                               & -              & -              & 95.02          & 95.61                                                         & -              & -                                                              \\
UniformFace\cite{uniformface} (CVPR2019)             & clean MS1M \cite{DBLP:conf/eccv/GuoZHHG16,deng2019arcface}   + VGGFace2  \cite{DBLP:conf/fgr/CaoSXPZ18}                                                                        & -              & -              & -              & -                                                              & 79.98          & 95.36                                                          \\
GroupFace\cite{groupface} (CVPR2020)                &  clean MS1M \cite{DBLP:conf/eccv/GuoZHHG16,deng2019arcface}                                                                      & 94.93          & 96.26          & 98.74 (3)          & 98.79                                                          & 81.31 (2)         &  97.35 (2)                                              \\
CircleLoss\cite{cricleloss} (CVPR2020)              & clean MS1M \cite{DBLP:conf/eccv/GuoZHHG16,cricleloss}                                                          & -              &  93.95          & 98.50          &  98.73                                                          & -              & -                                                              \\
CurricularFace\cite{curricularface} (CVPR2020)          & MS1MV2 \cite{DBLP:conf/eccv/GuoZHHG16,deng2019arcface}                                                                     & 94.8           & 96.1           & 98.71          & 98.64                                                          & 81.26 (3)         & 97.26                                                          \\
Dyn-arcFace \cite{dynarc} (MTAP2021)            & clean MS1M  \cite{DBLP:conf/eccv/GuoZHHG16,deng2019arcface}                                                                   & -              & -              & -              & -                                                              & -              & -                                                              \\
MagFace\cite{magface} (CVPR2021)                 & MS1MV2   \cite{DBLP:conf/eccv/GuoZHHG16,deng2019arcface}                                                                    & 94.51          & 95.97          & -              & -                                                              & -              & -                                                              \\
Partial-FC-ArcFace \cite{an2020partical_fc} (ICCVW2021)     & MS1MV2 \cite{DBLP:conf/eccv/GuoZHHG16,deng2019arcface}                                                                     & 94.8           & 96.2           & 98.31          & 98.59                                                          & -              & -                                                              \\
Partial-FC-CosFace \cite{an2020partical_fc} (ICCVW2021)    & MS1MV2 \cite{DBLP:conf/eccv/GuoZHHG16,deng2019arcface}                                                                     & 95.0             & 96.4           & 98.36          & 98.58                                                          & -              & -                                                              \\ \hline
ElasticFace-Arc (ours)   & MS1MV2  \cite{DBLP:conf/eccv/GuoZHHG16,deng2019arcface}                                                                    & 95.22  (3)        & 96.49 (3)         & \textbf{98.81} (1) & \textbf{98.92} (1)                                                & 80.76          &  97.30                                                          \\
ElasticFace-Cos (ours)   & MS1MV2 \cite{DBLP:conf/eccv/GuoZHHG16,deng2019arcface}                                                                     & 95.30 (2) & 96.57 (2) & 98.70           & 98.75                                                          & 81.01          & 97.31 (3)                                                         \\ 

ElasticFace-Arc+ (ours)   & MS1MV2  \cite{DBLP:conf/eccv/GuoZHHG16,deng2019arcface}                                                                    &95.09          & 96.40          & 98.80 (2) & 98.83  (3)                                              & 80.68          &  \textbf{97.44 }  (1)                                                      \\
ElasticFace-Cos+ (ours)   & MS1MV2 \cite{DBLP:conf/eccv/GuoZHHG16,deng2019arcface}                                                                     & \textbf{95.43} (1) & \textbf{96.65} (1) & 98.62           & 98.85   (2)                                                       & 80.08          & 97.29                                                          \\ \hline
\end{tabular}
}
\vspace{2mm}
\caption{The achieved results on the IJB-B, IJB-C, MegaFace (R), and MegaFace benchmarks. On the earlier three, and the verification accuracy of the fourth, the ElasticFace solutions consistently extend state-of-the-art performances. ElasticFace scores very close to the state-of-the-art on MegaFace. MegaFace has been refined in \cite{deng2019arcface} to MegaFace (R) as it contains many face images with wrong labels. All decimal points are provided as reported in the respective works. The top performance in each benchmark is in bold. The top three performances in each benchmark are noted with rank number between parentheses (1,2 or 3). }
\label{tab:res_ijb}
\end{table*}

\section{Results}
\label{sec:res}

Tables \ref{tab:res_lfw} and \ref{tab:res_ijb} presents the achieved results on the nine considered benchmarks. The main observation is that our proposed ElasticFace solutions scored beyond the state-of-the-art in seven out of the nine benchmarks, and very close to the state-of-the-art in the remaining two. When possible, and to build a fair comparison, the results of previous works are reported when trained on the MS1MV2 \cite{DBLP:conf/eccv/GuoZHHG16,deng2019arcface} (or a refined variant of MS1M \cite{DBLP:conf/eccv/GuoZHHG16}) as the ElasticFace results are based on training on this dataset.
The proposed ElasticFace ranked first in comparison to the state-of-the-art on the benchmarks AgeDB-30, CPLFW, CFP-FP, IJB-B, IJB-C, MegaFace (R), and MegaFace (verification). In the remaining benchmarks, ElasticFace solutions ranked second on CALFW, third on LFW, and fourth on MegaFace (identification).


A main outcome of the evaluation is concerning the databases with very large intra-user variations. These are the large age gape benchmark (AgeDB-30) and the frontal-to-profile face verification benchmark (CFP-FP). On AgeDB-30, our ElasticFace-Arc solution scored an accuracy of 98.35\%, while the top state-of-the-art performance was 98.32\% scored by the CurricularFace \cite{curricularface}.  
On CFP-FP, our ElasticFace-Arc+ solution scored an accuracy of 98.73\% and our ElasticFace-Arc scored an accuracy of 98.67\%, while the top state-of-the-art performances were 98.51\% scored by the Partial-FC-CosFace \cite{an2020partical_fc} 
solution and 98.46\% scored by the MagFace \cite{magface}. This significantly enhanced performance in the extreme intra-class variation scenarios points out the generalizability induced by the ElasticFace loss.
CALFW and CPLFW also considered age gaps and pose variation, however, with a lower variation than AgeDB-30 and CFP-FP. 
In CALFW, ElasticFace-Cos+ scored a close second with 96.18\% accuracy, with the lead going to the CurricularFace \cite{curricularface} with 96.20\% accuracy. In CPLFW, our ElasticFace-Arc+ is ranked first with 93.28\% accuracy, while the top state-of-the-art performance was 93.17\% accuracy scored by the GroupFace \cite{groupface}.
On the LFW benchmark \cite{LFWTech}, which is one of the oldest and nearly saturated benchmarks reported in the recent works, our ElasticFace-Cos and ElasticFace-Arc+ solutions scored an accuracy of 98.82\%, very close behind the GroupFace \cite{groupface} with 99.85\%. 

In Table \ref{tab:res_ijb}, on IJB-B benchmark, our ElasticFace-Cos+ scored a TAR at FAR1e–4 of 95.43\%, far ahead of the Partial-FC-CosFace \cite{an2020partical_fc} and the GroupFace \cite{groupface} with 95.0\% and 94.93\%, respectively. 
Similarly, on the IJB-C benchmark, our ElasticFace-Cos+ scored a TAR at FAR1e–4 of 96.65\%,  ahead of the Partial-FC-CosFace \cite{an2020partical_fc} and the GroupFace \cite{groupface} with 96.4\% and 96.36\% respectively. 
On the MegaFace (R), our ElasticFace-Arc scored 98.81\% Rank-1 identification rate and 98.92\% TAR at FAR1e–6, ahead of the previous lead solution, the GroupFace \cite{groupface} with 98.74\% and 98.79\%, respectively. 
On the MegaFace benchmark, our ElasticFace-Cos scored Rank-1 identification rate of 81.01\%, close to the state-of-the-art 82.72\% score by CosFace \cite{DBLP:conf/cvpr/WangWZJGZL018}, noting that CosFace was trained on a private dataset. On the same benchmark (MegaFace), our ElasticFace-Arc+ ranked first with 97.44\% TAR at FAR1e–6, while the top state-of-the-art performances were 97.35\% scored by the GroupFace \cite{groupface}.
It must be mentioned that the MegaFace benchmark has been refined in \cite{deng2019arcface} to MegaFace (R) as it contains many face images with wrong labels as reported in \cite{deng2019arcface}.

In comparison to the closely defined losses in ArcFace \cite{deng2019arcface}, CosFace \cite{DBLP:conf/cvpr/WangWZJGZL018}, and Partial-FC  \cite{an2020partical_fc} solutions, our ElasticFace models did prove to provide a strong performance edge by scoring higher recognition performance on most benchmarks.
When it comes to comparing ElasticFace and ElasticFace+, the ElasticFace-Arc and ElasticFace-Arc+ did achieve very close performances when considering all benchmarks.
On the other hand, the ElasticFace-Cos+ did outperform  ElasticFace-Cos on most benchmarks.

We acknowledge that the Partial-FC \cite{an2020partical_fc} solution reported additional performance rates when trained on their new collected database, the Glint360K \cite{an2020partical_fc}. However, we could not acquire this database as it requires an account on a cloud platform, that in itself requires a SIM card registered in a specific country,  which is very restrictive and we do not have access to. Therefore, and for a fair comparison, we opted to compare our results with the Partial-FC results when trained on the same dataset that our ElasticFace solution is using, the MS1MV2 \cite{DBLP:conf/eccv/GuoZHHG16,deng2019arcface} dataset.






The slightly increased training computational cost is a minor limitation of our proposed ElasticFace. Training the ResNet-100 model on MS1MV2 dataset with CosFace or ArcFace using the specified machine and training details described in Section \ref{sec:exp} requires around 57 hours.
This training time is increased by around one minute for ElasticFace and by 11 hours for ElasticFace+.
The minor increase in the ElasticFace training time is caused by the sampling of the margin values, while the larger increase in ElasticFace+ training time is additionally caused by the sorting algorithms.


On a less technical note, we stress that our efforts in the advancement of face recognition are aimed at enhancing the security, convenience, and life quality of the members of society,  e.g. enabling convenient access to financial and health services \cite{eadhaar} and enhancing the security of border checks within clear legal frameworks and users consent. We acknowledge and reject the possible malicious or illegal use of this and other technologies. 

\section{Conclusion}
In this paper, we propose an elastic margin penalty loss (ElasticFace) that avoids setting a single constant penalty margin.
Our motivation considers that real training data is inconsistent in terms of inter and intra-class variation, and thus the assumption made by many margin softmax losses that the geodesic distance between and within the different identities can be equally learned using a fixed margin is less than optimal.
We, therefore, relax this fixed margin constrain by using a random margin value drawn from a normal distribution in each training iteration.
In an extended definition, the assignment of these margin values to training samples corresponds to their proximity to their class centers. 
We evaluated our ElasticFace loss, in comparison to state-of-the-art face recognition approaches, on nine different benchmarks.
This evaluation demonstrated that our ElasticFace solution consistently extended state-of-the-art face recognition performance on most benchmarks (seven out of nine). This was specifically apparent in the challenging benchmarks with large intra-class variations, such as large age gaps and frontal-to-profile face comparisons. 
Our code, trained models, and training details will be released under the Attribution-NonCommercial-ShareAlike 4.0 International (CC BY-NC-SA 4.0) license.

{\small
\bibliographystyle{ieee}
\bibliography{main}
}

\end{document}